\documentclass[lettersize,journal]{IEEEtran}
\usepackage[T1]{fontenc}
\usepackage{amsmath,amsfonts}
\usepackage{algorithmic}
\usepackage{algorithm}
\usepackage{array}
\usepackage[caption=false,font=normalsize,labelfont=sf,textfont=sf]{subfig}
\usepackage{textcomp}
\usepackage{stfloats}
\usepackage{url}
\usepackage{verbatim}
\usepackage{graphicx}
\usepackage{cite}

\usepackage{hyperref} 
\usepackage{multirow}
\usepackage{booktabs}
\usepackage{xcolor}
\usepackage{marvosym}
\usepackage[numbers, square]{natbib}
\usepackage{adjustbox} 
\usepackage{longtable}
\usepackage{tikz}
\usepackage[edges]{forest}
\usetikzlibrary{shadows, calc, positioning}
\definecolor{hidden-draw}{RGB}{20,68,106}
\definecolor{hidden-pink}{RGB}{255,245,247}

\definecolor{hidden-black}{RGB}{20,68,106}
\definecolor{purple}{RGB}{144,153,196}
\definecolor{yellow}{RGB}{255,228,123}
\newcommand{\eg}{e.g.,\ }
\begin{document}

\title{A Comprehensive Survey of Synthetic Tabular Data Generation}

\author{Ruxue Shi, Yili Wang, Mengnan Du, Xu Shen, Yi Chang and Xin Wang\textsuperscript{(\Letter)}
 \thanks{Ruxue Shi, Yili Wang, Xu Shen, Yi Chang and Xin Wang\textsuperscript{(\Letter)} are with School of Artificial Intelligence, Jilin University, Changchun, Jilin 130012, China (Corresponding
authors: Xin Wang\textsuperscript{(\Letter)}, e-mail:\{shirx24, wangyl21, shenxu23\}@mails.jlu.edu.cn, \{yichang, xinwang\}@jlu.edu.cn).}
 \thanks{Mengnan Du is with the Department of Data Science, New Jersey Institute of Technology, Newark, USA (mengnan.du@njit.edu).}
}

\markboth{Journal of \LaTeX\ Class Files,~Vol.~14, No.~8, August~2021}%
{Shell \MakeLowercase{\textit{et al.}}: A Sample Article Using IEEEtran.cls for IEEE Journals}


\maketitle

\begin{abstract}

Tabular data is one of the most prevalent and important data formats in real-world applications such as healthcare, finance, and education. However, its effective use in machine learning is often constrained by data scarcity, privacy concerns, and class imbalance. Synthetic tabular data generation has emerged as a powerful solution, leveraging generative models to learn underlying data distributions and produce realistic, privacy-preserving samples. Although this area has seen growing attention, most existing surveys focus narrowly on specific methods (e.g., GANs or privacy-enhancing techniques), lacking a unified and comprehensive view that integrates recent advances such as diffusion models and large language models (LLMs).

In this survey, we present a structured and in-depth review of synthetic tabular data generation methods. Specifically, the survey is organized into three core components: (1) \textbf{Background}, which covers the overall generation pipeline, including problem definitions, synthetic tabular data generation methods, post processing, and evaluation; (2) \textbf{Generation Methods}, where we categorize existing approaches into traditional generation methods, diffusion model methods, and LLM-based methods, and compare them in terms of architecture, generation quality, and applicability; and (3) \textbf{Applications and Challenges}, which summarizes practical use cases, highlights common datasets, and discusses open challenges such as heterogeneity, data fidelity, and privacy protection.

This survey aims to provide researchers and practitioners with a holistic understanding of the field and to highlight key directions for future work in synthetic tabular data generation.

\end{abstract}

\begin{IEEEkeywords}
synthetic tabular data generation, generative methods, surveys.
\end{IEEEkeywords}

\section{Introduction}
\begin{figure*}[t]
\centering
\includegraphics[width=0.9\textwidth]{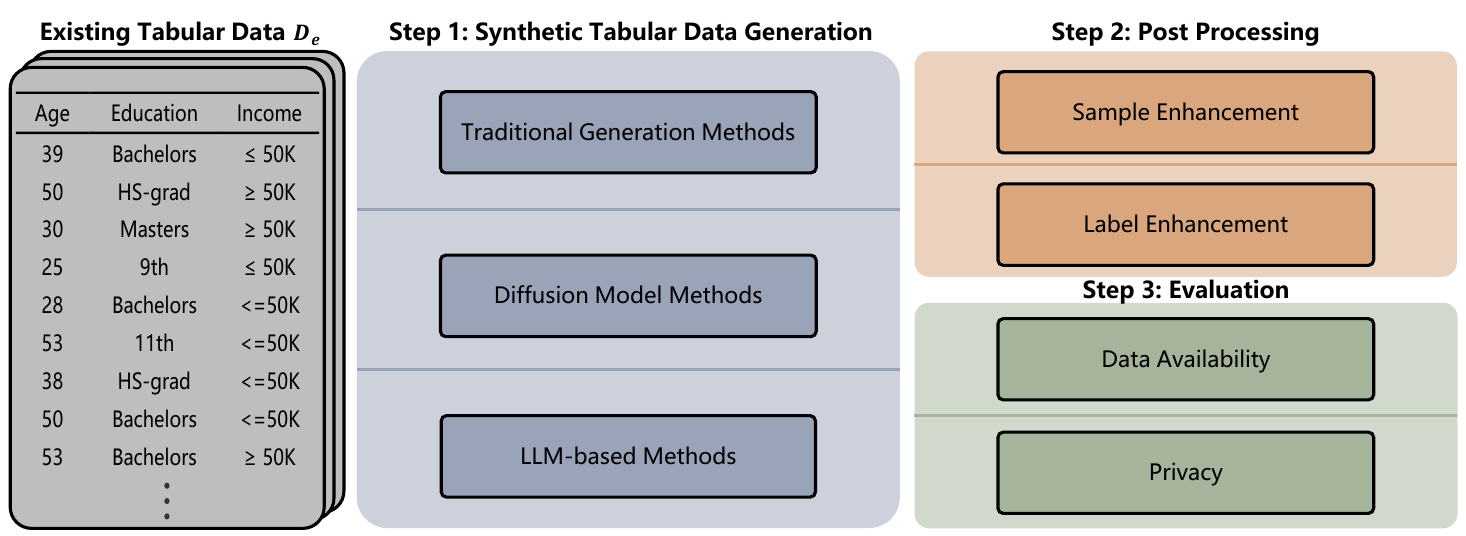}
\caption{\label{figure:pipeline}Pipeline of synthetic tabular data generation. The first step involves training generative models on existing tabular datasets to generate synthetic data. These models include Traditional Generation Methods, Diffusion Model Methods, and LLM-based Methods. To further improve the quality of the synthetic data and ensure alignment with human knowledge, the second step applies post-processing techniques, which include Sample Enhancement and Label Enhancement. Finally, in the third step, the synthetic data is evaluated in terms of Data Availability and Privacy.}
\end{figure*}
\IEEEPARstart{T}{abular} data is a fundamental data format in real-world applications, playing a crucial role in domains such as healthcare~\cite{shailaja2018machine}, finance~\cite{cao2022ai}, education~\cite{combrink2022comparing}, transportation~\cite{sun2022tabular}, and psychology~\cite{meehl1992theoretical}. Due to data privacy regulations such as the General Data Protection Regulation (GDPR)~\cite{regulation2018general} and California Consumer Privacy Act (CCPA)~\cite{illman2019california}, the data availability are restricted.
Furthermore, the inherent challenges of missing values and imbalanced class distributions in tabular data further limit its availability in practical applications. Fortunately, generative models offer a promising solution to these challenges by learning the distribution of tabular data and generating synthetic samples that resemble the tabular data while mitigating privacy concerns and data limitations.

Generative models such as Energy-based Models (EBMs)~\cite{lecun2006tutorial}, Variational Autoencoders (VAEs)~\cite{kingma2013auto}, Generative Adversarial Networks (GANs)~\cite{goodfellow2014generative}, Normalizing Flows~\cite{kobyzev2020normalizing}, Diffusion Models~\cite{sohl2015deep}, and Large Language Models (LLMs)~\cite{zhao2023survey} have been employed to synthesize tabular data with a distribution similar to the real data. 
Although existing generative methods have achieved remarkable performance on homogeneous data such as text and images, they often struggle to generate high-quality tabular data due to the inherent heterogeneity and complex distribution of tabular data. This challenge has attracted widespread interest and led to extensive research in this field. Such as TabDDPM~\cite{kotelnikov2023tabddpm}, TABDIFF~\cite{shi2024tabdiff} solve heterogeneity problems through multimodal diffusion. CLLM~\cite{seedat2023curated} leverages the internal knowledge of LLMs to improve generation quality in limited data scenarios. DP-LLMTGen~\cite{tran2024differentially}, DPGAN~\cite{xie2018differentially}, and PATE-GAN~\cite{jordon2018pate} enhance the privacy protection of both generative models and their outputs by incorporating differentially private strategies.

Although several existing studies have provided summaries of methods for tabular data generation, existing surveys have several limitations: (1) Some works are limited to specific domains or methods and lack a comprehensive overview of synthetic tabular data generation, such as, some surveys~\cite{hernandez2022synthetic, coutinho2021gans,umesh2024challenges} focus on the GANs for generating medical tabular data, while some surveys~\cite{hassan2023deep,panagiotou2024synthetic} focus on class imbalance, privacy, and fairness in synthetic tabular data generation. (2) Some works primarily focuses on the generation and evaluation of synthetic tabular data~\cite{kim2024generative,sauber2022use}, while overlooking the importance of post-processing after data generation. Synthetic data may match the original distribution but still violate commonsense or logical rules (e.g., negative age values or children marked as married). Therefore post-processing is necessary to ensure semantic correctness and data validity. (3) The latest advancements in diffusion models and LLMs for synthetic tabular data generation have not yet been thoroughly explored~\cite{fang2024large}. Diffusion models and LLMs offer advantages over traditional generative methods (e.g., diffusion models for their stability and representation capabilities, and LLMs for their ability to model semantic structure), but current surveys lack a thorough investigation of these methods. 

In this survey, we present a comprehensive review of synthetic tabular data generation, and the pipeline of synthetic tabular data generation is as shown in Figure~\ref{figure:pipeline}. We begin by formally defining the task of synthetic tabular data generation in order to clarify its objectives and the challenges.  Then, existing approaches are categorized into three categories: \textbf{traditional generation methods}, \textbf{diffusion model methods}, and \textbf{LLM-based methods}, and each category was analyzed in terms of its strengths and limitations.  In addition,  we summarize and classify the post-processing techniques into two primary categories: \textbf{sample enhancement} and \textbf{label enhancement}, both of which play a critical role in improving data quality and semantic validity. Evaluation methodologies are reviewed from the perspectives of data availability and privacy protection, reflecting the dual objectives of synthetic tabular data. Finally, we discuss the real-world applications and outline promising directions for future research.


The main contributions of this survey are summarized as follows:
\begin{itemize}
    \item \textbf{A deep comparative analysis across paradigms:} Building on an established three-fold taxonomy, we conduct a detailed, side-by-side comparison of traditional methods, diffusion models, and LLM-based approaches across multiple dimensions, including data type compatibility, sample fidelity, privacy resilience, and downstream performance.

    \item \textbf{The first systematic synthesis of post-processing strategies:} We introduce a novel taxonomy of post-processing methods for synthetic tabular data, distinguishing between sample enhancement and label enhancement techniques. This perspective bridges the gap between data generation and real-world deployment quality requirements.

    \item \textbf{An integrated evaluation framework:} We unify existing evaluation practices by organizing them into two principal dimensions, data availability (e.g., ML efficiency, fidelity, alignment) and privacy protection (e.g., membership inference attack, DCR), providing standardized guidance for future benchmarking.

    \item \textbf{A curated repository to promote reproducibility:} We compile and release a comprehensive collection of synthetic tabular data generation resources, including representative papers, open-source implementations, datasets, and evaluation tools, available at: https://github.com/ruxueshi/Awesome-Comprehensive-Survey-of-Synthetic-Tabular-Data-Generation.git.

\end{itemize}

\section{Definition}
To clearly describe the task of synthetic tabular data generation and its associated challenges, this section first provides a formal definition of the overall generation process. Then, we introduce the key challenges from three perspectives: data quality requirements, the inherent characteristics of tabular data, and the complexity of data distributions.
\subsection{Problem Definition}
Synthetic tabular data generation focuses on learning the statistical distribution of an tabular dataset \( D_e \) and producing a synthetic tabular dataset \( D_s \) that closely approximates it. Consequently, the overall generation task can be formulated as:

\begin{equation}
\mathcal{D}_{\text{s}} \gets p_\theta (\mathcal{D}_{\text{e}}),
\end{equation}

\noindent where \(p_\theta\) represents the generation model and \( \mathcal{D}_e = \left\{ (x_e^i, y_e^i) \right\}_{i=1}^{N_l} \) denotes a tabular dataset consisting of \( N_l \) samples, where \(x_e^i\) indicates all feature values of \(i\)-th row, and \(y_e^i\) represents its corresponding label. \( D_e \) and \( D_s \) follow the same structured format defined by a set of feature names \( F = \{ f_j \}_{j=1}^{d} \), where \( f_j \) represents the \( j \)-th feature name, and \( d \) denotes the total number of features. 

Since \( D_s \) often contains mislabeled and violate common human knowledge or domain-specific constraints samples, post-processing techniques are used to further improve the quality of the \( D_s \) and ensure the alignment with human knowledge, which include Sample Enhancement~\cite{seedat2023curated} and Label Enhancement~\cite{wang2021want}, see Section~\ref{sec:PP} for more details. Moreover, a comprehensive evaluation of data availability and privacy protection is also important to generate high-quality \( D_s \), in Section~\ref{sec:eval}, we summarize the evaluation metrics for data availability, including ML efficiency~\cite{fekri2019generating}, fidelity~\cite{scerbo2007high,hansen2023reimagining}, and alignment~\cite{stoian2025survey}, as well as the metrics for privacy protection~\cite{stoian2025survey}, such as Distance to Closest Record (DCR), Attribute Inference Attack, and Membership Inference Attack. These steps ensure that \( D_s \) can be effectively used in real-world applications while maintaining its privacy protection and data availability. 



\subsection{Challenges for Synthetic Tabular Data Generative Models}


\tikzstyle{my-box}=[
rectangle,
draw=hidden-black,
rounded corners,
text opacity=1,
minimum height=1.5em,
minimum width=5em,
inner sep=2pt,
align=center,
fill opacity=.5,
]
\tikzstyle{leaf}=[
my-box, 
minimum height=1.5em,
fill=yellow!32, 
text=black,
align=left,
font=\normalsize,
inner xsep=2pt,
inner ysep=4pt,
]
\tikzstyle{leaf2}=[
my-box, 
minimum height=1.5em,
fill=purple!27, 
text=black,
align=left,
font=\normalsize,
inner xsep=2pt,
inner ysep=4pt,
]
\tikzstyle{leaf3}=[
my-box, 
minimum height=1.5em,
fill=hidden-pink!90, 
text=black,
align=left,
font=\normalsize,
inner xsep=2pt,
inner ysep=4pt,
]

\begin{figure*}[t]
\vspace{-2mm}
\centering
\resizebox{\textwidth}{!}{
	\begin{forest}
		forked edges,
		for tree={
			grow=east,
			reversed=true,
			anchor=base west,
			parent anchor=east,
			child anchor=west,
			base=left,
			font=\large,
			rectangle,
			draw=hidden-black,
			rounded corners,
			align=left,
			minimum width=4em,
			edge+={darkgray, line width=1pt},
			s sep=3pt,
			inner xsep=2pt,
			inner ysep=3pt,
			line width=0.8pt,
			ver/.style={rotate=90, child anchor=north, parent anchor=south, anchor=center},
		},
		where level=1{text width=10em,font=\normalsize,}{},
		where level=2{text width=8.0em,font=\normalsize,}{},
		where level=3{text width=9.5em,font=\normalsize,}{},
		where level=4{text width=12em,font=\normalsize,}{},
		[Synthetic Tabular Data Generation,ver [Traditional Generation \\ Methods~(\S\ref{sub:traditional})
			[Machine Learning  \\Method
                    [\eg~Copulas~\cite{sklar1973random}{,} Gaussian Mixture Models~\cite{reynolds2009gaussian}{,} SMOTE~\cite{chawla2002smote}{,} Borderline-\\SMOTE~\cite{han2005borderline}{,} SMOTE-CDNN~\cite{wang2023synthetic}{,} SMOTE-ENC~\cite{mukherjee2021smote}{,} ADASYN~\cite{he2008adasyn}{,} Synthpop~\cite{nowok2016synthpop}, leaf, text width=42em
                    ]   
                ]
			[VAE-based \\Method
                    [\eg~GOGGLE~\cite{liu2023goggle}{,} VAEM~\cite{ma2020vaem}{,} TVAE~\cite{xu2019modeling}, leaf, text width=42em
                    ]
                ]
                [GAN-based \\Method
                [Traditional GAN, text width=8em
                    [\eg~TGAN~\cite{xu2018synthesizing}{,} medGAN~\cite{choi2017generating}{,} CasTGAN~\cite{alshantti2024castgan}{,} CrGAN-Cnet~\cite{mottini2018airline}{,}  Table-\\GAN~\cite{park2018data}{,} IT-GAN~\cite{lee2021invertible}{,} VT-GAN~\cite{zhao2023gtv}{,} FCT-GAN~\cite{zhao2022fct}{,} TabFairGAN~\cite{rajabi2022tabfairgan}, leaf, text width=32.2em
                    ]
                ]
                [Conditional GAN, text width=8em
                [\eg~cGANS~\cite{douzas2018effective}{,} CTAB-GAN~\cite{zhao2021ctab}{,} OCT-GAN~\cite{kim2021oct}{,} cWGAN~\cite{engelmann2020conditional}{,} \\ CTGAN~\cite{xu2019modeling}{,} RC-TGAN~\cite{gueye2023row}, leaf, text width=32.2em
                    ]
                ]
                [DP GAN, text width=8em
                [\eg~DPGAN~\cite{xie2018differentially}{,} PATE-GAN~\cite{jordon2018pate}{,} RDP-CGAN~\cite{torfi2022differentially}{,} CTAB-GAN+~\cite{zhao2024ctab}, leaf, text width=32.2em
                    ]
                ]
                ]
		]
		[Diffusion Model \\Methods~(\S\ref{subsec:diffusion})
			[DDPM-based \\Method
                    [Generic, text width=8em [\eg~TabDDPM~\cite{kotelnikov2023tabddpm}{,} TABDIFF~\cite{shi2024tabdiff}{,} CoDi~\cite{lee2023codi}{,} CTSyn~\cite{lin2024ctsyn}{,} \\AutoDiff~\cite{suh2023autodiff}{,} RelDDPM~\cite{liu2024controllable}, leaf2, text width=32.2em]
                    ]
                    [Domain-specific, text width=8em
                        [\eg~TabDDPM-HER~\cite{ceritli2023synthesizing}{,} FinDiff~\cite{sattarov2023findiff}{,} DPM-HER~\cite{nicholas2023synthetic}{,} Imb-FinDiff~\cite{schreyer2024imb}{,} \\FLEXGEN-EHR~\cite{he2024flexible}{,} EntTabDiff~\cite{liu2024entity}, leaf2, text width=32.2em]
                    ]
                ]
			[Score-based \\Method
                    [\eg~SOS~\cite{kim2022sos}{,} MissDiff~\cite{ouyang2023missdiff}{,} CSDI\_T~\cite{zheng2022diffusion}{,} STaSY~\cite{kim2022stasy}{,} Forest-VP/Forest Flow~\cite{jolicoeur2024generating}{,} TABSYN~\cite{zhang2023mixed}, leaf2, text width=42em]
                ]
			]
            [LLM-based \\Methods~(\S\ref{subsec:llm})
                [Prompt-based \\Method
                    [\eg~EPIC~\cite{kim2024epic}{,} CLLM~\cite{seedat2023curated}{,} LITO~\cite{yang2024language}{,} OCTree~\cite{nam2024optimized}, leaf3, text width=42em]
                ]
			[Fine-Tuning \\Method
                [\eg~GReaT~\cite{borisov2022language}{,} TabMT~\cite{gulati2023tabmt}{,} TAPTAP~\cite{zhang2023generative}{,} HARMONIC~\cite{wang2024harmonic}{,} AIGT~\cite{zhang2024aigt}{,} DPLLMTGen~\cite{tran2024differentially}{,} \\Pred-LLM~\cite{nguyen2024generating}{,} REaLTabFormer~\cite{realtabformer}{,} TabuLa~\cite{zhao2023tabula}{,} P-TA~\cite{yang2024p}, leaf3, text width=42em]
                ]
			]
		]
        \end{forest}
        }
\vspace{-6mm}
\caption{Overview of synthetic tabular data generation methods, including \textit{Traditional Generation Methods}, \textit{Diffusion Model Methods} and \textit{LLM-based Methods}.}
\label{fig:taxonomy}
\vspace{-3mm}
\end{figure*}
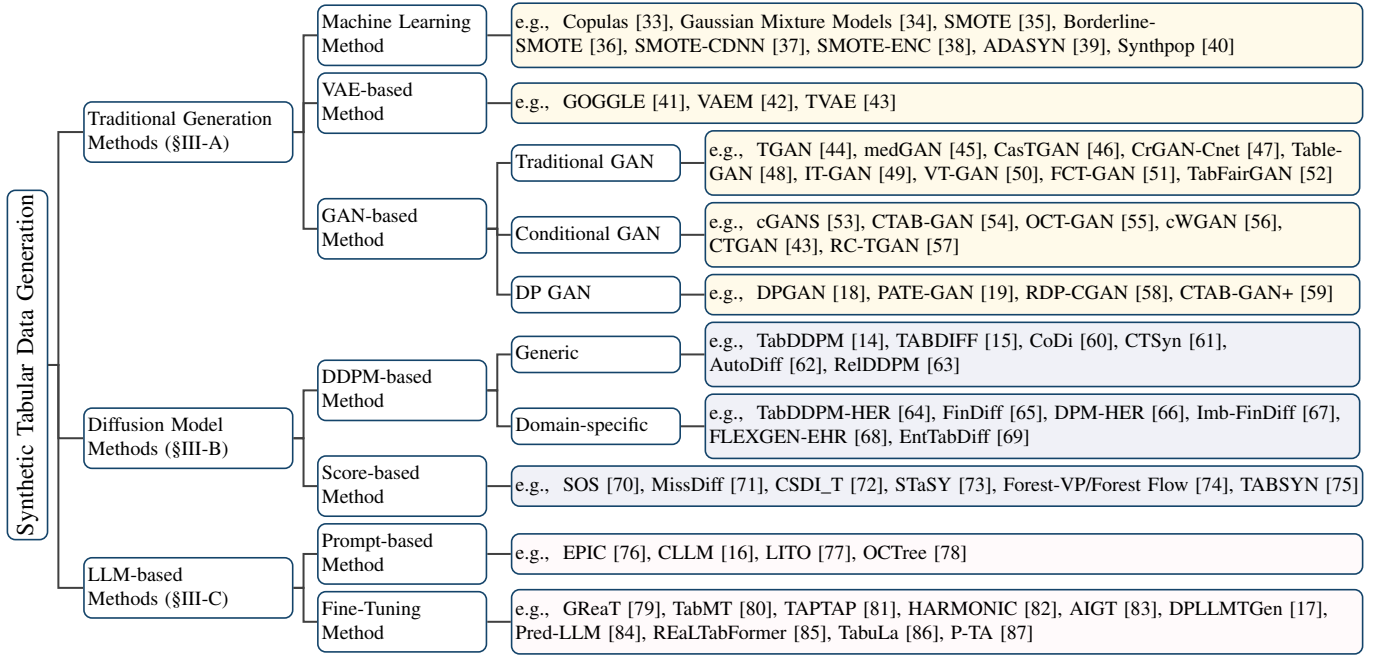

Synthetic tabular data is more challenges unlike synthetic text and image data, which consist of uniform basic units (e.g., pixels or tokens) that allow the generative model \(p_\theta\) to learn distributions in a homogeneous space, tabular data is inherently heterogeneous and structurally diverse, posing substantial challenges for both effective modeling and the generation of high-quality synthetic data. In the following, we present the key challenges of synthetic tabular data generation from three perspectives: data quality, inherent characteristics of tabular data, and complex data distributions.

\subsubsection{Data Quality} The quality issues of existing tabular dataset \( D_e \) bring the following challenges:
\begin{itemize}
    \item \textbf{Limited Data Size.} Tabular data, particularly in specialized domains such as healthcare, finance, and agriculture, is often challenging to collect and annotate. As a result, most datasets are limited in size, which significantly restricts their availability. This data scarcity poses challenges for training generative models.
    \item \textbf{Imbalanced Class Distribution.}  Tabular data frequently exhibits severe class imbalance, where the data size of certain categories is significantly lower. For instance, in medical datasets, the number of patients is often much lower than that of healthy individuals. Similarly, in manufacturing, defective products make up only a small fraction of the data. Such an imbalance makes it difficult for generative models to learn the representations of minority classes, often resulting in biased synthetic outputs.
    \item \textbf{Missing Values.} Missing values are common in tabular datasets due to privacy protection constraints, human error, or the optional nature of certain attributes. For example, in electronic health records, not every diagnostic test is performed for every patient, leading to incomplete entries. These missing values increase modeling complexity, as generative models must learn underlying distributions while handling incomplete observations.  

\end{itemize}

\subsubsection{Inherent Characteristics} The inherent characteristics of tabular data pose the following challenges for current methods:
\begin{itemize}
   \item \textbf{Heterogeneity.} Unlike homogeneous data types (e.g., pixels or tokens), tabular data contains a mix of categorical and numerical columns, each with distinct statistical properties. This heterogeneity makes it difficult for conventional generative models to model the joint distribution of such mixed data effectively.

   \item \textbf{Complex and Sparse Inter-column Dependencies.} Tabular data often contains complex inter-column dependencies, including both homogeneous (e.g., numerical-numerical) and heterogeneous (e.g., categorical-numerical) dependencies. However, these relationships are typically sparse—only certain column pairs may be correlated. For instance, age and education level might be interrelated, whereas others remain independent. Accurately capturing these sparse and complex dependencies is crucial to preserving realism in synthetic data \( D_s \).

    \item \textbf{Mixed-Type Single Column.} In some cases, a column of tabular data may contain both numerical and categorical data. This further complicates modeling, as it challenges traditional encoding strategies and increases the difficulty of learning coherent representations within a single column.

\end{itemize}

\subsubsection{Complex data Distributions} The complex distribution in tabular data bring additional challenges to the generation of high-quality synthetic data \( D_s \).
\begin{itemize}

   \item \textbf{Non-Gaussian Distribution of Numerical Columns.} Numerical columns in tabular data are often different from Gaussian distributions, exhibiting skewness, heavy tails, or multimodal characteristics. This poses a challenge for existing generative models to learn such complex distributions. 

   \item \textbf{Imbalanced Categories in Categorical Columns.} Categorical columns often exhibit imbalanced distributions, where certain categories dominate the feature. Generative models may overfit these dominant classes, leading to poor representation of minority classes and reducing the diversity of \( D_s \).

   \item \textbf{Marginal Effects of Columns.} The same values may have different meanings depending on the column name. For example, the value “10” could represent age in age column, or rating in rate column, depending on the which column it belongs to. Maintaining these semantic distinctions is essential to preserving the interpretability and correctness of \( D_s \).
\end{itemize}

\section{Synthetic Tabular Data Generation Methods}
\label{sec:method}
In response to the emergence of various generation approaches, we propose a novel taxonomy that categorizes existing methods into three categories: Traditional Generation Methods, Diffusion Model Methods, and LLM-based Methods. For each category, we introduce representative methods, analyze their respective strengths and limitations, and provide a comprehensive and reliable support for future research in this field.
\subsection{Traditional Generation Methods}
\label{sub:traditional}
\begin{table*}[t]
  \centering
  \caption{Overview of traditional generation methods for synthetic tabular data generation.}
    \begin{tabular}{clrll}
    \toprule
    \multicolumn{2}{c}{Model} & \multicolumn{1}{c}{Year} & Venue & Model Architecture \\
    \midrule
    \multirow{7}[2]{*}{Machine Learning Method} & Copulas~\cite{sklar1973random} & \multicolumn{1}{c}{1973} & Kybernetika & Statistical models \\
          & Synthpop~\cite{nowok2016synthpop} & \multicolumn{1}{c}{2016} & JSS   & Statistical models \\
          & SMOTE~\cite{chawla2002smote} & \multicolumn{1}{c}{2002} & JAIR  & Distance-based \\
          & \multicolumn{1}{p{10.165em}}{Borderline-SMOTE~\cite{han2005borderline}} & \multicolumn{1}{c}{2005} & ICIC  & Distance-based \\
          & ADASYN~\cite{he2008adasyn} & \multicolumn{1}{c}{2008} & IJCNN & Distance-based \\
          & SMOTE-ENC~\cite{mukherjee2021smote} & \multicolumn{1}{c}{2021} & Applied System Innovation & Distance-based \\
          & SMOTE-CDNN~\cite{wang2023synthetic} & \multicolumn{1}{c}{2023} & Applied Soft Computing & Distance-based \\
    \midrule
    \multirow{3}[2]{*}{VAE-based Method} & TVAE~\cite{xu2019modeling}  & \multicolumn{1}{c}{2019} & NeurIPS & Simple VAE \\
          & VAEM~\cite{ma2020vaem}  & \multicolumn{1}{c}{2020} & NeurIPS & Multiple VAEs \\
          & GOGGLE~\cite{liu2023goggle} & \multicolumn{1}{c}{2023} & ICLR  & Graph Neural Network \\
    \midrule
    \multirow{19}[2]{*}{GAN-based Method} & medGAN~\cite{choi2017generating} & 2017  & MLHC  & Traditional GAN \\
          & TGAN~\cite{xu2018synthesizing}  & 2018  & arXiv & Traditional GAN \\
          & CrGAN-Cnet~\cite{mottini2018airline} & 2018  & arXiv & Traditional GAN \\
          & TableGAN~\cite{park2018data} & 2018  & PVLDB & Traditional GAN \\
          & IT-GAN~\cite{lee2021invertible} & 2021  & NeurIPS & Traditional GAN \\
          & FCT-GAN~\cite{zhao2022fct} & 2022  & arXiv & Traditional GAN \\
          & TabFairGAN~\cite{rajabi2022tabfairgan} & 2022  & MAKE  & Traditional GAN \\
          & VT-GAN~\cite{zhao2023gtv} & 2023  & arXiv & Traditional GAN \\
          & CasTGAN~\cite{alshantti2024castgan} & 2024  & IEEE Access & Traditional GAN \\
          & cGANS~\cite{douzas2018effective} & 2018  & EXPERT SYST APPL & Conditional GAN \\
          & CTGAN~\cite{xu2019modeling} & 2019  & NeurIPS & Conditional GAN \\
          & cWGAN~\cite{engelmann2020conditional} & 2020  & EXPERT SYST APPL & Conditional GAN \\
          & CTAB-GAN~\cite{zhao2021ctab} & 2021  & ACML  & Conditional GAN \\
          & OCT-GAN~\cite{kim2021oct} & 2021  & WWW   & Conditional GAN \\
          & RC-TGAN~\cite{gueye2023row} & 2023  & ICASSP & Conditional GAN \\
          & DPGAN~\cite{xie2018differentially} & 2018  & arXiv & Differential Privacy GAN \\
          & PATE-GAN~\cite{jordon2018pate} & 2019  & ICLR  & Differential Privacy GAN \\
          & RDP-CGAN~\cite{torfi2022differentially} & 2022  & Information Sciences & Differential Privacy GAN \\
          & CTAB-GAN+~\cite{zhao2024ctab} & 2024  & Frontiers Big Data & Differential Privacy GAN \\
    \bottomrule
    \end{tabular}%
  \label{tab:traditional}%
\end{table*}%

\begin{figure*}[t]
\centering
\includegraphics[width=0.7\textwidth]{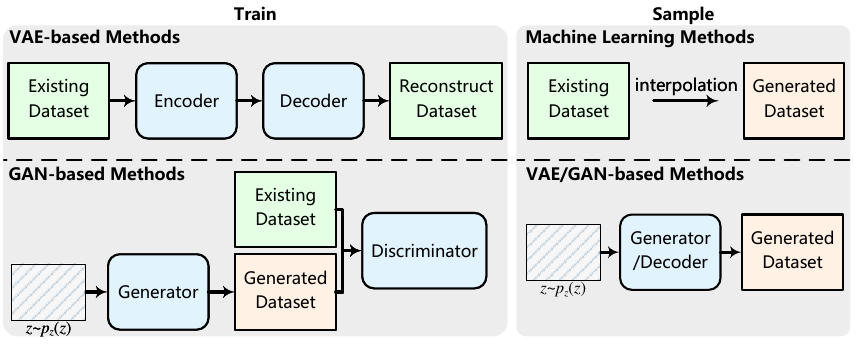}
\caption{\label{figure:TGM}Traditional generation methods, including \textit{machine learning methods}, \textit{VAE-based Methods} and \textit{GAN-based Methods}.}
\end{figure*}

In the early stages of synthetic tabular data generation, traditional methods such as machine learning methods, VAE-based methods, and GAN-based methods, as shown in Figure~\ref{figure:TGM} are the main approaches for synthetic tabular data generation. Among them, GAN-based methods are further categorized into three categories based on their application scenarios: Traditional GANs, Conditional GANs, and Differentially Private GANs, as summarized in Table~\ref{tab:traditional}.  

\subsubsection{Machine Learning-based Methods} Before the specially designed generative models such as Variational Autoencoders (VAEs), and Generative Adversarial Networks (GANs), machine learning (ML) techniques such as Copulas~\cite{sklar1973random}, Gaussian Mixture Models (GMMs)~\cite{reynolds2009gaussian}, and Bayesian networks~\cite{rabaey2024clinical} were widely employed for data generation. These methods rely on explicit parameterization and estimation of joint probability distributions. Additionally, specialized resampling methods were developed, including shallow interpolation-based techniques such as the Synthetic Minority Over-sampling Technique (SMOTE)~\cite{chawla2002smote} and its variants (e.g., Borderline-SMOTE~\cite{han2005borderline}, SMOTETomek~\cite{wang2023synthetic}, and SMOTE-ENC~\cite{mukherjee2021smote}), as well as Adaptive Synthetic Sampling (ADASYN)~\cite{he2008adasyn} and probabilistic methods like Synthpop~\cite{nowok2016synthpop} for data synthesis and interpolation. However, these models struggle with capturing complex data distributions, particularly in high-dimensional settings where accurately learning feature dependencies becomes challenging. Meanwhile, their reliance on predefined assumptions and explicit probability estimation often limits their effectiveness in modeling intricate relationships present in real-world tabular data. 

\subsubsection{VAE-based Method} 
With the rise of deep learning, generative models specifically designed for tabular data synthesis have gained prominence. Compared to ML learning techniques, deep learning-based generative models, such as VAE, further improve the quality of synthetic tabular data by learning latent representations by maximizing the Evidence Lower Bound (ELBO)~\cite{alemi2018fixing}, which is defined as:
\begin{equation}
\log p(x) \geq \mathbb{E}_{q(z|x)}[\log p(x|z)] - KL(q(z|x)\,\|\,p(z)),
\end{equation}
\noindent where \( q(z|x) \) is the encoder network, \( p(x|z) \) is the decoder network, and \( KL \) denotes the Kullback–Leibler divergence.

TVAE~\cite{xu2019modeling} adapts VAEs to tabular data by transforming categorical features into numerical ones using one-hot encoding. VAEM~\cite{ma2020vaem} is trained in a two-stage manner, the first stage provides a more uniform representation of the data to the second stage, thereby avoiding the problems caused by heterogeneous data. GOGGLE~\cite{liu2023goggle} introduces Graph Neural Networks (GNN)~\cite{wu2020comprehensive} into the encoder and decoder of VAE, and learns a graph adjacency matrix to capture inter-column dependencies. Although VAE-based Methods have been widely used in the field of synthetic tabular generation, the generated tends towards averaging and loss of details.
\subsubsection{GAN-based Generators} The objective of GANs~\cite{goodfellow2020generative} is formulated as a classic min-max game between a generator and a discriminator:
\begin{equation}
    \min_G \max_D \; \mathbb{E}_{x \sim p_{\text{data}}(x)}[\log D(x)] + \mathbb{E}_{z \sim p_z(z)}[\log(1 - D(G(z)))],
\end{equation}
\noindent where the generator \( G \) aims to produce a synthetic sample \(x_s\) that can fool the discriminator \( D \), while the discriminator seeks to distinguish real data \(x_e\) from generated samples. GAN-based methods for tabular data generation are typically categorized into three groups: Traditional GAN, Conditional GAN, and Differentially Private GAN.

\textbf{Traditional GAN.} Several studies have extended GANs for tabular data by modifying the model architecture to accommodate categorical features. TGAN~\cite{xu2018synthesizing} utilizes Long Short-Term Memory (LSTM)~\cite{graves2012long} with attention mechanisms~\cite{vaswani2017attention} to generate both numerical and categorical features one by one. MedGAN~\cite{choi2017generating} develops an autoencoder-based generative model for generating high-dimensional electronic health records (EHR). CasTGAN~\cite{alshantti2024castgan} is a generative network framework characterized by multiple generators connected sequentially, where each generator is responsible for generating a feature, and one discriminator validates the output of all the generators. CrGAN-Cnet~\cite{mottini2018airline} employs GANs for generating airline passenger name records, incorporating Cramer distance~\cite{bellemare2017cramer} and a Cross-Net architecture to improve the model's handling of missing values. TableGAN~\cite{park2018data} uses Convolutional Neural Networks (CNNs)~\cite{lecun1989backpropagation} for both the generator and discriminator, when the tabular data includes a label column, a prediction loss is added to the generator to better improve the correlation between the label column and the other columns. IT-GAN~\cite{lee2021invertible} proposes a generalized GAN framework for synthetic tabular data generation, combining adversarial training with negative logarithmic density regularization in reversible neural networks, and balances the generation quality with privacy protection leakage. VT-GAN~\cite{zhao2023gtv} introduces a training-with-shuffling technique to prevent any party from reconstructing the training data from the GAN's conditional vector, enabling distributed tabular data synthesis in a privacy protection manner. FCT-GAN~\cite{zhao2022fct} incorporates feature tokenization and Fourier networks to construct a transformer-style generator and discriminator, capturing both local and global inter-column dependencies. TabFairGAN~\cite{rajabi2022tabfairgan} uses a two-phase training process: the first phase generates synthetic data similar to the reference dataset, and the second phase modifies the value function to add fairness constraints, ensuring that the synthetic data is both accurate and fair. While these methods enhance synthetic tabular data generation, they lack the ability to conditionally generate data based on specific feature constraints, such as generating health records for female patients. 


\textbf{Conditional GAN.} To address the limitations of traditional GAN in controlling the features of synthetic data \(\mathcal{D}_{\text{s}}\), Conditional GAN~\cite{miyato2018cgans} have been widely adopted,which is an extension of the GAN framework. An additional conditional space \(C\) is introduced, which represents the external information coming from the training data. The Conditional GAN framework modifies the model \(G\) and \(D\), to include the additional conditional space \(C\), as follows:
\begin{equation}
\begin{split}
\min_G \max_D V(D, G) &=  \; \mathbb{E}_{x, c \sim p_{\text{data}}(x, c)}[\log D(x, c)] \\
& + \mathbb{E}_{z \sim p_z(z), c \sim p_y(c)}[\log(1 - D(G(z, c), c))],
\end{split}
\end{equation}
\noindent where \( (x, c) \in X \times C \) are sampled from the data distribution \( p_{\text{data}}(x, c) \).


cGANs~\cite{douzas2018effective} incorporates a conditional vector, allowing the generation of data for specific classes, which is particularly useful when dealing with highly imbalanced datasets. This capability is crucial for applications such as initializing datasets in online learning scenarios~\cite{hoi2021online}. CTAB-GAN~\cite{zhao2021ctab} designs a novel conditional vector that efficiently encodes mixed data types and addresses skewed distributions, mitigating data imbalance and long-tail challenges. OCT-GAN~\cite{kim2021oct} designs the generator and discriminator based on Neural Ordinary Differential Equations (NODEs)~\cite{pinckaers2019neural}, demonstrating that NODEs have theoretically advantageous properties for synthetic tabular data generation. OCT-GAN’s generator includes an ODE layer at the beginning of its architecture to transform the initial input vector into a latent space suitable for the generation process. cWGAN~\cite{engelmann2020conditional} integrates Wasserstein distance~\cite{panaretos2019statistical} into the Conditional GAN framework, leveraging conditional vectors to oversample minority classes for more balanced data generation. CTGAN~\cite{xu2019modeling} enhances cGANs~\cite{douzas2018effective} by incorporating the PacGAN~\cite{lin2018pacgan} structure in the discriminator, employing a combination of generator loss, Wasserstein loss with gradient penalty, and a training-by-sampling strategy to handle imbalanced categorical features. RC-TGAN~\cite{gueye2023row} is proposed to handle complex multi-tabular datasets by conditioning the generation of the child tabular data on the previously generated parent rows, specifically using the features of the parent. This method transfers relationship information from the parent tabular data to the child tabular data.


\textbf{Differentially Private GAN.} To ensure differential privacy protection during training, the differentially private stochastic gradient descent (DP-SGD)~\cite{xie2018differentially} algorithm is widely employed. The core idea is to add calibrated noise to the averaged and clipped per-sample gradients in each iteration. The modified gradient used for updating the model parameters at iteration \( t \) is defined as follows:
\begin{equation}
\tilde{g}_t = \frac{1}{L} \left( \sum_{i=1}^{L} \text{clip}(g_t^i, C) \right) + \mathcal{N}(0, \sigma^2 C^2 \mathbf{I}),
\end{equation}
\noindent where \( g_t^i \) denotes the gradient computed on the \( i \)-th training sample in a minibatch of size \( L \). Each per-sample gradient is individually clipped using an \(\ell_2\)-norm threshold \( C \) to ensure bounded sensitivity:
\begin{equation}
    \text{clip}(g_t^i, C) = g_t^i \cdot \min\left(1, \frac{C}{\|g_t^i\|_2}\right).
\end{equation}
After clipping, the gradients are averaged and additive Gaussian noise \( \mathcal{N}(0, \sigma^2 C^2 \mathbf{I}) \) is injected to achieve differential privacy protection. The parameter \( \sigma \) controls the noise scale, balancing the trade-off between privacy protection and model utility. The identity matrix \( \mathbf{I} \) ensures isotropic noise across dimensions. This mechanism ensures that the influence of any single training sample on the model update is limited, thereby making it impossible for attackers to infer specific individual information even if they obtain model output.


DPGAN~\cite{xie2018differentially} achieves differential privacy protection by adding carefully designed noise to the gradients during the learning process. PATE-GAN~\cite{jordon2018pate} utilizes the Private Aggregation of Teacher Ensembles (PATE)~\cite{papernot2018scalable} framework, perturbing the output of multiple teacher discriminators using Laplacian noise~\cite{sarathy2011evaluating} before training a student discriminator. However, a key limitation of PATE-GAN~\cite{jordon2018pate} is that the student discriminator only learns from synthetic data, which may lead to unreliable feedback. RDP-CGAN~\cite{torfi2022differentially} builds upon convolutional autoencoders and generative adversarial networks, incorporating Rényi Differential Privacy (RDP)~\cite{mironov2017renyi} to provide tighter bounds on privacy costs and improve the privacy-utility trade-off. CTAB-GAN+~\cite{zhao2024ctab} builds upon existing methods by incorporating RDP-based privacy accounting, similar to DP-WGAN~\cite{huang2022dpwgan}. Furthermore, by leveraging the Was+GP loss, CTAB-GAN+ effectively constrains the gradient norm, eliminating the need for weight clipping and resulting in more stable training for differentially private GANs.


\subsection{Diffusion Model Methods}
\label{subsec:diffusion}
\begin{table*}[t]
  \centering
  \caption{Overview of diffusion models for synthetic tabular data generation.}
    \begin{tabular}{clccl}
    \toprule
    \multicolumn{2}{c}{Model} & Year  & Latent Space & Venue \\
    \midrule
    \multirow{12}[2]{*}{DDPM-based} & TabDDPM~\cite{kotelnikov2023tabddpm} & 2023  & No  & ICML \\
          & CoDi~\cite{lee2023codi}  & 2023  & No  & ICML \\
          & AutoDiff~\cite{suh2023autodiff} & 2023  & Yes & SyntheticData4ML \\
          & RelDDPM~\cite{liu2024controllable} & 2024  & No & Proc. ACM Manag. Data \\
          & CTSyn~\cite{lin2024ctsyn} & 2025  & Yes & ICLR \\
          & TABDIFF~\cite{shi2024tabdiff} & 2025  & No & ICLR  \\
          & TabDDPM-HER~\cite{ceritli2023synthesizing} & 2023  & No & arXiv  \\
          & DPM-HER~\cite{nicholas2023synthetic} & 2023  & No & SyntheticData4ML \\
          & FLEXGEN-EHR~\cite{he2024flexible} & 2024  & Yes & ICLR  \\
          & EntTabDiff~\cite{liu2024entity} & 2023  & Yes & ICAIF \\
          & FinDiff~\cite{sattarov2023findiff} & 2024  & No & ICAIF \\
          & Imb-FinDiff~\cite{schreyer2024imb} & 2024  & Yes & ICAIF \\
    \midrule
    \multirow{6}[2]{*}{Score-based} & SOS~\cite{kim2022sos}   & 2022 & No & KDD \\
          & MissDiff~\cite{ouyang2023missdiff} & 2022  & No & TRL @ NeurIPS \\
          & CSDI\_T~\cite{zheng2022diffusion} & 2023  & No & SPIGM @ ICML \\
          & STaSY~\cite{kim2022stasy} & 2023 & No & ICLR \\
          & Forest-VP/Forest-Flow~\cite{jolicoeur2024generating} & 2024  & No & PMLR \\
          & TABSYN~\cite{zhang2023mixed} & 2024  & Yes & ICLR \\
    \bottomrule
    \end{tabular}%
  \label{tab:diff}%
\end{table*}%

\begin{figure}[t]
\centering
\includegraphics[width=0.5\textwidth]{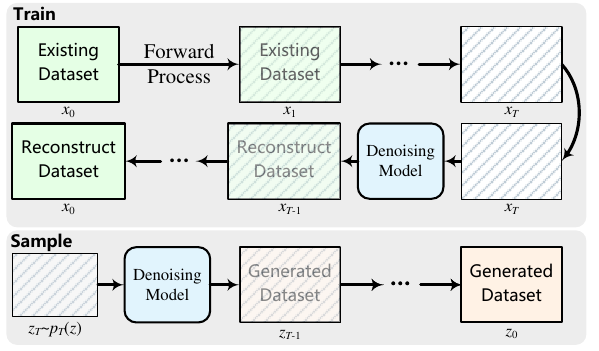}
\caption{\label{figure:diff}The Training and sample of Diffusion Model Methods.}
\end{figure}



Although GANs have made notable progress in synthetic tabular data generation, these models often encounter challenges, including training instability, mode collapse, and poor representation of multimodal distributions, which hinder the generation of high-quality synthetic tabular data. In contrast, diffusion models effectively address these challenges through their specialized denoising mechanism, which gains significant research interest. The training and sampling process of diffusion model methods is shown in Figure~\ref{figure:diff}. In the domain of synthetic tabular data generation, diffusion models have begun to exhibit advantages similar to those of GANs and VAEs, achieving notable performance improvements. These models demonstrate strong potential in overcoming the unique challenges associated with complex tabular data distributions modeling, highlighting their effectiveness in generating high-quality, diverse synthetic data. The current mainstream tabular diffusion generation models include \textit{DDPM-based Method}, and \textit{Score-based Method} as shown in Table~\ref{tab:diff}.  

\subsubsection{DDPM-based Method}  Diffusion models~\cite{sohl2015deep,ho2020denoising} are likelihood-based generative models that synthesize data by learning to reverse a gradual noising process, which model data generation as a Markovian process consisting of two stages: a forward (diffusion) process, which progressively adds noise to data, and a reverse (denoising) process, which reconstructs data from noise. Given an initial data point \( x_0 \sim q(x_0) \), the forward process defines a series of latent features \( x_1, x_2, \ldots, x_T \) by iteratively adding noise sampling from the predefined distributions \(q(x_t | x_{t-1})\) with variances \(\{\beta_1, ..., \beta_T\}\):
\begin{equation}
    q(x_{1:T} | x_0) = \prod_{t=1}^{T} q(x_t | x_{t-1}),
\end{equation}
\noindent the goal of the reverse process is to learn the denoising transitions \( p_\theta(x_{t-1} \mid x_t) \) to recover data from noise:
\begin{equation}
   p_\theta(x_{0:T}) = \prod_{t=1}^{T} p_\theta(x_{t-1} | x_t),
\end{equation}
where the parameters \( \theta \) are optimized by maximizing a variational lower bound on the data log-likelihood:

\begin{equation}
\begin{split}
\log q(x_0) \geq & 
\\ \mathbb{E}_{q(x_{0:T})} \Big[ 
    & \log p_\theta(x_0 \mid x_1) 
    - \mathrm{KL}(q(x_T \mid x_0) \,\|\, q(x_T)) \\
    & - \sum_{t=2}^{T} \mathrm{KL}(q(x_{t-1} \mid x_t, x_0) \,\|\, p_\theta(x_{t-1} \mid x_t)) 
\Big].
\end{split}
\end{equation}

Due to the inherent heterogeneity of tabular data, existing Denoising Diffusion Probabilistic Models (DDPMs) face challenges in adapting to tabular data. To address this, researchers modify DDPMs to better accommodate the structural characteristics of tabular data. TabDDPM~\cite{kotelnikov2023tabddpm} extends DDPM to tabular data and demonstrates its superiority over existing GAN/VAE alternatives, which is consistent with the advantages of diffusion models in other fields. TABDIFF~\cite{shi2024tabdiff} develops a joint continuous-time diffusion process for numerical and categorical data and proposes feature-wise learnable diffusion processes to counter the diversity of different feature distributions. CoDi~\cite{lee2023codi} uses two diffusion models to handle numerical and categorical features, respectively. During the training process, these two diffusion models mutually read conditions and constrain each other, solving the problem of modeling categorical features. CTSyn~\cite{lin2024ctsyn} develops a unified aggregator that tokenizes and embeds heterogeneous tabular rows to a unified latent space, and facilitates the training of models across tabular data. AutoDiff~\cite{suh2023autodiff} leverages autoencoders to encode tabular data into a latent space, where the diffusion process is applied. This approach effectively addresses challenges associated with heterogeneous features, mixed data types, and complex inter-column dependencies, enabling more accurate and structured synthetic tabular data generation. RelDDPM~\cite{liu2024controllable} leverages diffusion models to first learn an unconditional generative model. Subsequently, RelDDPM~\cite{liu2024controllable} introduced lightweight controllers to guide the unconditional generative model in generating synthetic data that satisfies different conditions.

In addition to employing DDPM-based methods for general-purpose synthetic tabular data generation, several studies have explored their application in specific domains—most notably, \textit{healthcare}. For instance, TabDDPM-EHR~\cite{ceritli2023synthesizing} applies TabDDPM to the generation of electronic health records (EHR), demonstrating its ability to synthesize high-quality EHR data while preserving the statistical properties of the original dataset. DPM-EHR~\cite{nicholas2023synthetic} leverages DDPM to generate EHR data consisting of mixed-type features, including numerical, binary, and categorical features. Furthermore, FLEXGEN-EHR~\cite{he2024flexible} introduces an optimal transformation module that aligns heterogeneous EHRs into a common feature space, while effectively addressing the promplem of missing values within a unified learning framework. In the \textit{finance} domain, diffusion-based models have also shown promising results. EntTabDiff~\cite{liu2024entity} generates synthetic entities by learning the distribution of entity-level features and subsequently synthesizes tabular data conditioned on these entities. FinDiff~\cite{sattarov2023findiff} is designed to produce high-fidelity synthetic financial tabular data suitable for various downstream tasks. Building upon this, Imb-FinDiff~\cite{schreyer2024imb} further addresses the challenge of class imbalance by focusing on the generation of samples from minority categories within long-tail distributions, thereby enhancing the representativeness and utility of synthetic financial datasets.

\subsubsection{Score-based Method} Score-based method learns a score function, which represents the gradient of the logarithmic density of the data distribution. Unlike traditional diffusion models that rely on variational bounds and discrete-time Markov chains, score-based models use the gradient of the log density score function to define and reverse a continuous-time diffusion trajectory. Let \( x \) be a continuous-time feature that follows forward SDE~\cite{song2020score,song2020improved}:
\begin{equation}
    dx = f(x, t)\,dt + g(t)\,dw,
\end{equation}
\noindent where \( f(x, t) \) is the drift coefficient, \( g(t) \) is the diffusion coefficient (typically scalar), and \( w \) is the standard Wiener process. In the forward process, noise is gradually added to the data. The reverse-time SDE is given by:
\begin{equation}
    dx = \left[f(x, t) - g(t)^2 \nabla_x \log p_t(x)\right] dt + g(t)\, d\overline{w},
\end{equation}
\noindent where \( \nabla_x \log p_t(x) \) is the score function of the data distribution at time \( t \), and \( d\overline{w} \) is the reverse-time Wiener process. Since \( \log p_t(x) \) is unknown, it is estimated using a score network \( s_\theta(x, t) \approx \nabla_x \log p_t(x) \), trained using denoising score matching:
\begin{equation}
\begin{split}
\mathcal{L}(\theta) = 
& \; \mathbb{E}_{t \sim \mathcal{U}(0, T)} \lambda(t)
\, \mathbb{E}_{x_0, x_t} \big[
\| s_\theta(x_t, t) \\
& \; - \nabla_x \log p_t(x_t \mid x_0) \|_2^2
\Big]
\end{split}
\end{equation}
\noindent where \( \lambda(t) \) is a weighting function, \( x_0 \sim p_{\text{data}} \), and \( x_t \) is the noisy version of \( x_0 \). Once trained, the model generates samples by solving the reverse SDE (e.g., using numerical solvers like Euler-Maruyama~\cite{mao2015truncated} or Predictor-Corrector schemes~\cite{gragg1964generalized}), starting from Gaussian noise \( x_T \sim \mathcal{N}(0, I) \) and integrating backward in time. These models have gained attention for their flexibility and effectiveness in capturing complex data distributions, making them well-suited for generating high-fidelity synthetic data across various domains. SOS~\cite{kim2022sos} is the first score-based tabular data oversampling method, which redesigns the network to process tabular data, and through fine-tuning, the oversampling quality has been further improved. MissDiff~\cite{ouyang2023missdiff} introduces a unified diffusion-based framework that incorporates a masked regression loss for denoising score matching during the training phase. This approach enables the model to effectively learn from data with missing values, addressing various missing data problems and improving the quality of synthetic data in the presence of incomplete information. To effectively process both categorical and numerical features simultaneously, CSDI-T~\cite{zheng2022diffusion} explores three techniques: one-hot encoding, analog bit encoding, and feature tagging. STaSy~\cite{kim2022stasy} leverages self-paced learning techniques and fine-tuning strategies to enhance the stability of denoising score-matching training, leading to improved sampling quality and greater diversity in synthetic tabular data. Forest-VP/Forest-Flow~\cite{jolicoeur2024generating} employs XGBoost~\cite{chen2016xgboost}, leveraging its strength in tabular data prediction and classification. Importantly, XGBoost~\cite{chen2016xgboost} can naturally handle missing data by learning the best split, allowing the model to be trained directly on incomplete datasets—an advantage over most generative models that require complete data. TABSYN~\cite{zhang2023mixed} synthesizes tabular data by leveraging a diffusion model within a variational autoencoder (VAE) crafted latent space.

\subsection{LLM-based Methods}
\label{subsec:llm}
\begin{figure*}[t]
\centering
\includegraphics[width=0.8\textwidth]{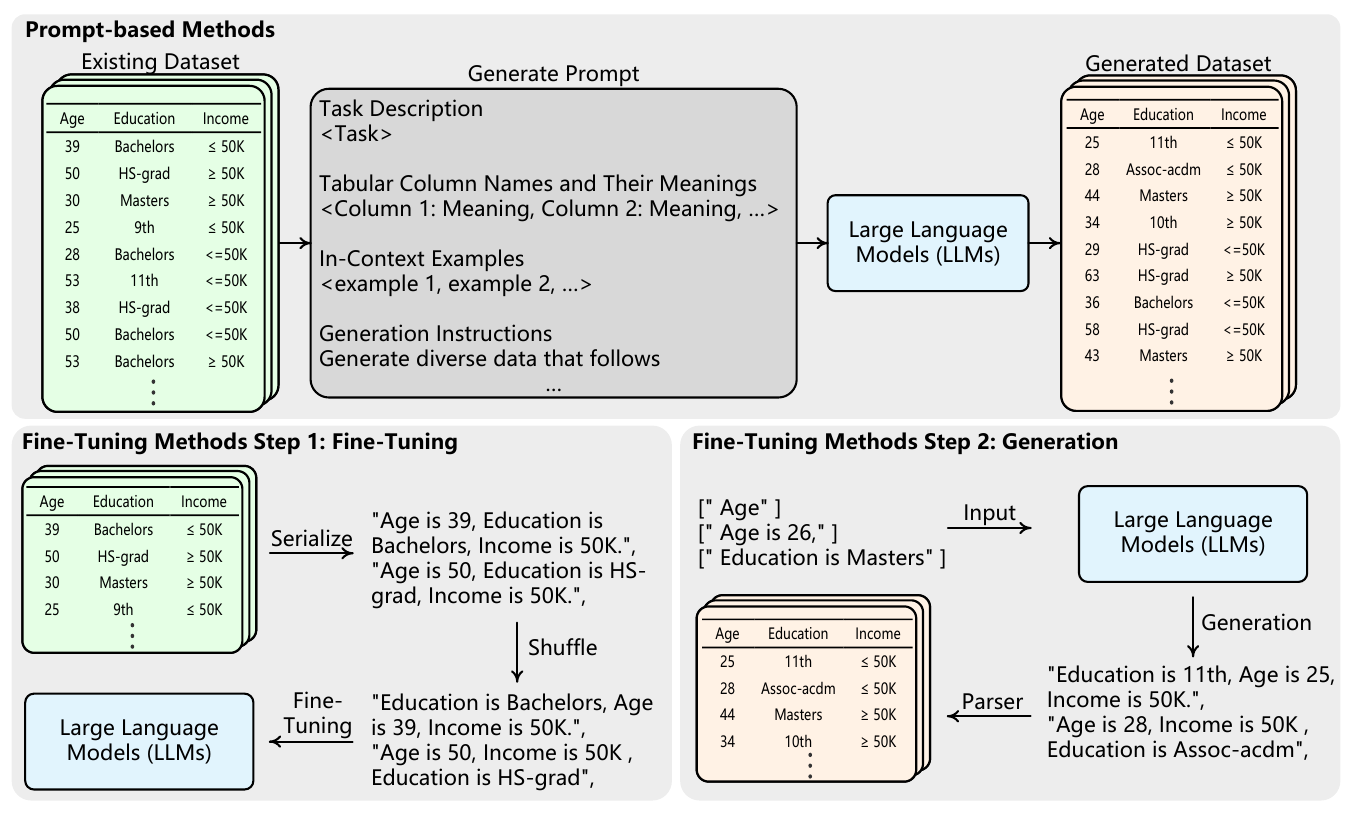}
\caption{\label{figure:LLMs}LLM-based methods, including \textit{Prompt-based Method} and \textit{Fine-Tuning Methods}.}
\end{figure*}

\begin{table*}[t]
  \centering
      \caption{Overview of LLMs-based models for synthetic tabular data generation.}
  \resizebox{0.8\textwidth}{!}{
  
    \begin{tabular}{clclll}
    \toprule
    \multicolumn{2}{c}{Model} & Year  & Venue & Serialization & Used Model \\
    \midrule
    \multirow{3}[2]{*}{Prompt-based} & EPIC~\cite{kim2024epic}  & 2024  & NeurIPS & CSV   & GPT-3.5,Mistral-7b-v0.1,Llama-2-7b \\
          & CLLM~\cite{seedat2023curated}  & 2024  & ICML & Text  & GPT-4,GPT-3.5 \\
          & LITO~\cite{yang2024language}  & 2024  & ICLR  & Text  & Distill-GPT2,GPT3.5 \\
    \midrule
    \multirow{8}[2]{*}{Fine-Tuning}  & GReaT~\cite{borisov2022language} & 2023  & ICLR  & Text  & Distill-GPT2,GPT2 \\
    & TabuLa~\cite{zhao2023tabula} & 2023  & arXiv & Text  & Distill-GPT2 \\
    
          & TabMT~\cite{gulati2023tabmt} & 2023  & NeurIPS & Mask  & BERT \\
          & TAPTAP~\cite{zhang2023generative} & 2023  & EMNLP & Text  & Distill-GPT2,GPT2 \\
          & HARMONIC~\cite{wang2024harmonic} & 2024  & NeurIPS & Text  & LLaMA-2-7b-chat \\
          
          & DP-LLMTGen~\cite{tran2024differentially} & 2024  & arXiv & Text  & LLaMA-2-7b-chat \\
          & Pred-LLM~\cite{nguyen2024generating} & 2024  & arXiv & Text  & Distill-GPT2 \\
          & P-TA~\cite{yang2024p}  & 2024  & ACL   & Text  & GPT-2,GPT-Neo \\
          & AIGT~\cite{zhang2024aigt}  & 2025  & ACL   & Text  & DistilGPT-2,Llama3.1-8B \\
    \bottomrule
    \end{tabular}%
    }

  \label{tab:LLMs}%
\end{table*}%
Generating high-quality and realistic tabular data, particularly in specialized domains such as finance and healthcare, often requires domain-specific knowledge. For instance, generating synthetic heart disease records necessitates an understanding that the probability of heart disease increases with age, alongside general knowledge, such as the fact that a person's age cannot be negative. However, existing generative models typically focus on learning the statistical distribution of raw tabular data without incorporating such knowledge. As a result, they may generate data that aligns with the learned distribution but violates logical or domain-specific constraints. To address this challenge, recent research explores leveraging the knowledge embedded in extensively pre-trained LLMs for synthetic tabular data generation. An overview of LLM-based models for synthetic tabular data generation is summarized in Table~\ref{tab:LLMs}. Depending on whether the LLM is fine-tuned, these methods can be broadly categorized into \textit{Prompt-based} methods and \textit{Fine-Tuning} methods as shown in Figure~\ref{figure:LLMs}.

\subsubsection{Prompt-based Methods} 
In-context learning (ICL) offers an innovative approach by enabling language models to perform tasks based solely on input-output examples, without parameter updates or fine-tuning. This capability, first observed in GPT-3 and subsequent LLMs~\cite{liu2023pre}, allows models to generalize to new tasks by embedding examples directly into the input prompt, which is often referred to as an "emergent ability"~\cite{wei2022emergent}, this phenomenon has spurred significant research into understanding and enhancing ICL in models exceeding 100 billion parameters. Leveraging this capability, existing methods serialize the tabular dataset \( D_e \) into a format compatible with LLM's input. By carefully designing generation prompts, these methods guide the LLM to synthesize a synthetic tabular dataset \( D_s \). EPIC~\cite{kim2024epic} leverages balanced, grouped data samples and consistent formatting with unique feature mapping to guide LLMs in generating accurate synthetic data across all classes, even for imbalanced datasets. CLLM~\cite{seedat2023curated} leverages the prior knowledge of LLMs for data augmentation in the limited data scenarios, and introduces a principled curation mechanism, leveraging learning dynamics, coupled with confidence and uncertainty metrics, to obtain a high-quality dataset. LITO~\cite{yang2024language} proposes an oversampling framework for tabular data to guide the abilities of generative language models. By leveraging its conditional sampling capabilities, LITO~\cite{yang2024language} synthesizes minority samples by progressively masking the important features of the majority class samples and imputing them towards the minority distribution. 

\subsubsection{Fine-Tuning Methods} 
Directly prompting LLMs for synthetic tabular data generation may lead to inaccurate or misleading outputs that deviate from user instructions. To mitigate this issue, recent research has explored fine-tuning LLMs on tabular data, enabling them to better understand the structure, constraints, and relationships within tabular datasets. Fine-tuning helps improve the accuracy and coherence of synthetic tabular data, ensuring that the output aligns more closely with real-world data distributions and domain-specific knowledge. GReaT~\cite{borisov2022language} fine-tunes LLM for sampling synthesized but highly realistic tabular data. In addition, GReaT can model the distribution of tabular data by adjusting any subset of features. TabMT~\cite{gulati2023tabmt} is a Bert-based mask transformer design used to generate synthetic tabular data, effectively addressing the unique challenges posed by heterogeneous data fields and capable of handling missing value problem. TAPTAP~\cite{zhang2023generative} utilizes tabular data pre-training to enhance the model's tabular prediction ability. After pre-training with a large amount of real-world tabular data, TAPTAP~\cite{zhang2023generative} can generate high-quality tabular data to support various applications, including privacy protection, limited data scenarios, missing value imputation, and imbalanced classification. HARMONIC~\cite{wang2024harmonic} constructs an instruction fine-tuning dataset based on the idea of the k-nearest neighbors algorithm to inspire LLMs to discover inter-row relationships. By fine-tuning, LLMs are trained to remember the format and connections of the data rather than the data itself, which reduces the risk of privacy leakage. AIGT~\cite{zhang2024aigt} based on prompt enhancement, which utilizes metadata information, such as tabular descriptions and schemas, as prompts to generate a synthetic tabular dataset  \( D_s \) and propose long token partitioning algorithms that enable AIGT to model tabular data of any scale. DP-LLMTGen~\cite{tran2024differentially} for differentially private tabular data synthesis that leverages pre-trained large language models (LLMs) to model sensitive datasets using a two-stage fine-tuning procedure with a novel loss function specifically designed for tabular data. Subsequently, it generates synthetic data through sampling the fine-tuned LLMs. P-TA~\cite{yang2024p} proposes the use of Proximal Policy Optimization (PPO~\cite{schulman2017proximal}) to apply GANs to guide LLMs to refine the probability distribution of tabular features and thereby generate more realistic data.

\section{Post Processing}
\label{sec:PP}
\begin{figure*}[t]
\centering
\includegraphics[width=0.9\textwidth]{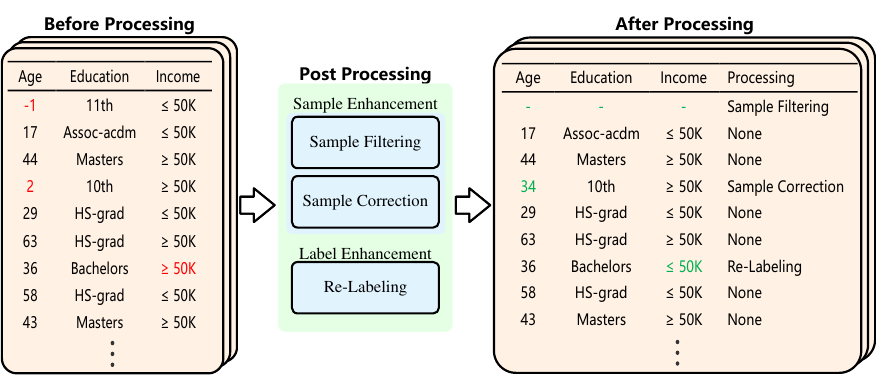}
\caption{\label{figure:post}Overview of post processing.}
\end{figure*}
The generative model introduced in Section~\ref{sec:method} is theoretically capable of generating an infinite amount of tabular data. However, these synthetic samples often contain a number of unrealistic, incorrect, or even toxic instances. This issue arises from two main factors. First, the distribution of tabular data is highly complex, making its generation a traditionally challenging task. Second, non-LLM-based methods lack common sense and professional knowledge, leading to the generation of data that conforms to the statistical distribution but deviates from human knowledge. In LLM-based tabular generation methods, the "hallucination problem" exacerbates this issue, as LLMs inevitably produce flawed samples with incorrect labels. To address these challenges, several post-processing techniques for synthetic tabular data have been proposed. These methods can be broadly categorized into two groups: \textit{sample enhancement} and \textit{label enhancement}, as shown in Figure~\ref{figure:post}.
\subsection{Sample Enhancement}
Sample enhancement methods aim to improve the quality and rationality of synthetic tabular data by modifying feature values in the synthetic samples or filtering the unreasonable sample to transform originally unreasonable or invalid samples into more realistic ones. They can be broadly categorized into two main approaches: sample filtering and sample correction.

\subsubsection{Sample Filtering} The core of sample filtering methods lies in identifying low-quality samples. Therefore, the key step is to design appropriate criteria based on learning dynamics~\cite{arpit2017closer, arora2019fine, li2020learning}, which may include factors such as confidence scores, impact functions, and generation capabilities, then samples are filtered according to these established criteria.
For example, CLLM~\cite{seedat2023curated} analyzes learning dynamics to determine whether to sample or not by calculating two key metrics: confidence and uncertainty. Samples that are uncertain or produce inconsistent predictions can harm model performance, so they are filtered out. Similarly, LITO~\cite{yang2024language} uses self-authentication with rejection sampling. The language model itself can validate the correctness of synthetic samples by inferring labels and filtering out inaccurate or inconsistent ones, improving the reliability of the data without external validation.

\subsubsection{Sample Correction} Sample correction incorporates neuro-symbolic principles into the training process of tabular data generation models by introducing human knowledge in the form of constraints or rules. Such as, LL~\cite{stoian2024realistic} addresses the issue of aligning synthetic samples with background knowledge by adding an additional layer to restrict the output space to comply with a set of linear inequalities, ensuring consistency between synthetic data and available domain knowledge. However, this approach is limited to cases where the knowledge can be accurately expressed using linear inequalities, whose representational capacity is inherently constrained. To overcome this limitation, DRL~\cite{stoian2025beyond} extends the approach by representing background knowledge using Quantifier-Free Linear Real Arithmetic (QFLRA) formulas, allowing the model to enforce constraints in non-convex and even disconnected spaces.

\subsection{Label Enhancement}
Label enhancement methods aim to correct potential annotation errors in synthetic samples. The most straightforward approach is manual re-annotation of mislabeled data. Wang et al.~\cite{wang2021want} propose actively selecting samples with the lowest confidence scores for human re-labeling, while Pangakis et al.~\cite{pangakis2023automated} and Liu et al.~\cite{liu2022wanli} further emphasize the importance of manual review in ensuring data quality. However, despite its effectiveness, manual re-labeling incurs significant costs and is often impractical for large-scale deployment. To address these limitations, most synthetic tabular data generation methods adopt the proxy model approach, which leverages automated models to refine and correct labels. TAPTAP~\cite{zhang2023generative} and AIGT~\cite{zhang2024aigt} use generative models to generate tabular features and then use the backbone discriminative model trained on the original data to generate pseudo labels for the synthetic data. Pred-LLM~\cite{nguyen2024generating} generates tabular features through pre-training LLM, while further prompting LLM to generate its labels.

\section{Evaluation}
\label{sec:eval}
\begin{table*}[t]
  \centering
  \caption{Overview of popular tabular data generation datasets. Only datasets that have been used by more than five relevant methods are included in this table. Abbr stands for abbreviation of dataset name.}
    \begin{tabular}{clccccll}
    \toprule
    Data Source & Dataset & Abbr  & \# Rows & \# Columns (Num/Cat) & Miss Values & Task Type & Domain \\
    \midrule
    \multirow{8}[1]{*}{UCI} & Abalone & AB    & 4177  & 8(7/1) & None  & Regression & Biology \\
          & Bank  & BA    & 45211 & 16(6/10) & None  & Binclass & Business \\
        & Default & DE    & 30000 & 23(20/3) & None  & Binclass & Business \\
          & News  & NE    & 39797 & 59(45/14) & None  & Regression & Business \\
          & Beijing & BE    & 43824 & 11(2/9) & Yes   & Regression & Climate and Environment \\
          & Obesity & OB    & 2111  & 17(8/9) & None  & Multiclass & Healthcare \\
          & Shoppers & SH    & 12330 & 17(9/8) & None  & Binclass & Marketing \\ 
          & Magic & MA    & 19020 & 10(10/0) & None  & Binclass & Physics and Chemistry \\
    \midrule
    \multirow{6}[0]{*}{OpenML}           & Covertype & CO    & 423680 & 54(10/44) & None  & Binclass & Biology \\
    & Credit & CG    & 1000  & 20(8/12) & None  & Binclass & Finance \\
    & Diabetes & DI    & 768   & 8(8/0) & None  & Binclass & Healthcare \\
    & Sick  & SI    & 3711  & 27(6/21) & None  & Binclass & Healthcare \\ 
    & Car   & CR    & 1728  & 6(6/0) & None  & Multiclass & Marketing \\
    & Adult & AD    & 48842 & 14(6/8) & Yes   & Binclass & Social Science \\
\midrule
    \multirow{6}[1]{*}{Kaggle} 
     & Churn & CH    & 10000 & 11(7/4) & None  & Binclass & Finance \\
    & Heloc & HE    & 9872  & 23(23/0) & None  & Binclass & Finance \\
    & Loan  & LO    & 5000  & 13(6/7) & None  & Binclass & Finance \\
    & Insurance & IN    & 1338  & 6(3/3) & None  & Regression & Healthcare \\
    & California Housing & CA    & 20640 & 8(8/0) & Yes   & Regression & Marketing \\
    & King  & KI    & 21613 & 20(17/3) & None  & Regression & Marketing \\

    \bottomrule
    \end{tabular}%
  \label{tab:dataset}%
\end{table*}%

The performance of tabular data generative models is typically evaluated from multiple, complementary perspectives. In this section, we categorize the evaluation of synthetic tabular data generation into two dimensions: \textbf{Data Availability} and \textbf{Privacy Protection}. \textit{Data availability} refers to the effectiveness of the synthetic data in supporting downstream machine learning tasks, including learning efficiency, fidelity to the original data distribution, and alignment with human domain knowledge. \textit{Privacy Protection}, on the other hand, measures the extent to which the synthetic data avoids disclosing sensitive information that could potentially be used to re-identify user information from the original dataset.

\subsection{Dataset}
For the convenience of subsequent research in synthetic tabular data generation, we present a summary of widely used benchmark datasets in the field of synthetic tabular data generation. Commonly used tabular datasets for synthetic data generation are primarily sourced from three major repositories: UCI Machine Learning Repository, OpenML, and Kaggle. Most existing studies on tabular data generation rely on datasets from these platforms. In this survey, we summarize 20 widely used datasets across various tasks, including regression, binary classification, and multi-class classification. Among them, datasets such as Beijing, Adult, and California Housing are frequently used in studies focusing on synthetic tabular data generation with missing values, as shown in Table~\ref{tab:dataset}.
\subsection{Data Availability}
To address the challenges of insufficient data, class imbalance, and missing values in real-world tabular datasets, data augmentation has become a central objective of synthetic tabular data generation models. The effectiveness of the synthetic data for this target is measured through its utility, encompassing three key metrics: ML efficiency (i.e., how well the synthetic data supports downstream learning tasks), fidelity (i.e., how closely it matches the statistical properties of the original data), and alignment (i.e., how well it conforms to domain-specific human knowledge).

\subsubsection{ML Efficiency} Machine Learning (ML) utility is a key criterion for evaluating the effectiveness of synthetic tabular data, as it reflects the extent to which the synthetic data can support downstream learning tasks. One widely adopted evaluation protocol for this purpose is the Training on Synthetic and Testing on Real (TSTR) scheme~\cite{fekri2019generating}. In this framework, tabular data is first divided into a training set and a test set. A generative model is then trained on the training data and used to produce a synthetic tabular dataset \( D_s \) of the same size. Subsequently, two machine learning models are trained separately: one on the real training data and the other on the synthetic data. Both models are then evaluated using the same real test set. The core assumption behind this scheme is that if the synthetic data accurately captures the underlying patterns and distributions of the original data, then a model trained on the synthetic data should generalize well to real-world data. Therefore, if the performance of the model trained on synthetic data is comparable to that of the model trained on real data (e.g., in terms of accuracy, F1 score, AUC, and RMSE, etc.), the synthetic tabular dataset \( D_s \) is deemed to possess high machine learning utility. notably \(\text{AUC} = \int_{0}^{1} \text{ROC}(t) \, d(t)\) is commonly used for evaluating categorical data, while \(\text{RMSE} = \sqrt{ \frac{1}{n} \sum_{i=1}^{n} (y_i - \hat{y}_i)^2 }\) is commonly used for regression datasets.

\subsubsection{Fidelity} Fidelity measures how well synthetic data preserves the statistical properties of the tabular data. Although recent studies~\cite{scerbo2007high,hansen2023reimagining} suggest that high fidelity does not always imply strong downstream task performance, it remains a crucial aspect of evaluating synthetic data quality.

Fidelity evaluation methods can be broadly categorized into three types: \begin{itemize} \item 



\textbf{Column-wise:} This metrics compares individual features between real and synthetic data. Common metrics include the Kolmogorov-Smirnov Test (KST)~\cite{berger2014kolmogorov} for numerical features. Given two numerical distributions, $p_r(x)$ (representing real data) and $p_s(x)$ (representing synthetic data), KST quantifies the distance between these distributions by calculating the maximum discrepancy between their corresponding Cumulative Distribution Functions (CDFs):

\begin{equation}
\text{KST} = \sup_x \left| F_r(x) - F_s(x) \right|,
\end{equation}

where $F_r(x)$ and $F_s(x)$ are the CDFs of $p_r(x)$ and $p_s(x)$, respectively:

\begin{equation}
F(x) = \int_{-\infty}^x p(x) \, dx.
\end{equation}

For categorical data, the Total Variation Distance (TVD)~\cite{tao2024discriminative} is commonly used. TVD computes the frequency of each category and expresses it as a probability. The TVD score represents the average difference between the probabilities of each category:

\begin{equation}
\text{TVD} = \frac{1}{2} \sum_{\omega \in \Omega} \left| R(\omega) - S(\omega) \right|, 
\end{equation}

where $\omega$ represents all possible categories in a given column $\Omega$, and $R(\cdot)$ and $S(\cdot)$ denote the real and synthetic frequencies for these categories, respectively.

\item \textbf{Pair-wise:} Assesses how well dependencies between feature pairs are preserved using metrics such as Pearson Correlation Coefficient Matrices (DPCM) (for numerical features), The difference between Contingency Similarity Matrices among categorical features (DCSM) (for categorical features), and The difference between Contingency Similarity Matrices (for numerical and categorical features). \item \textbf{Joint distribution:} Evaluates the similarity between the full joint distributions of real and synthetic data, such as \text{$\beta$-Recall.} and Coverage score. \end{itemize}

While fidelity alone is not sufficient to determine data utility, it remains a fundamental metric for assessing the structural realism of synthetic tabular data.
\subsubsection{Alignment} Alignment with background knowledge has recently attracted growing attention from the research community and is increasingly recognized as a critical criterion for evaluating the authenticity of synthetic data. There are three primary metrics commonly used to evaluate alignment in synthetic tabular data: (1) Constraint Violation Rate (CVR), which measures the percentage of synthetic samples that violate at least one predefined constraint; (2) Constraint Violation Coverage (CVC), which calculates the proportion of constraints that are violated by at least one sample in the dataset; and (3) Sample-wise Constraint Violation Coverage (sCVC), which computes the average percentage of samples that violate each individual constraint.

\begin{table*}[htbp]
  \centering
  \caption{The performance of various methods for synthesizing tabular data is evaluated using multiple metrics. AUC and RMSE are used to assess the ML Efficiency of the synthesized data on classification and regression tasks, respectively. CVR evaluates how well the synthetic data aligns with background knowledge. Column-wise and Pair-wise metrics are employed to measure the fidelity of the synthesized data. The best results are highlighted in bold.}
  \begin{tabular}{clccccccc}
    \toprule
    \multirow{2}[4]{*}{Data} & \multirow{2}[4]{*}{Metrics} & \multirow{2}[4]{*}{Real} & \multicolumn{6}{c}{Method} \\
    \cmidrule{4-9}
          &       &   & CTGAN & CTGAN+ & TVAE  & P-TA  & TabDDPM & TABSYN \\
    \midrule
    \multirow{4}[2]{*}{Adult}& $\uparrow$AUC   & 92.50\(_{0.25}\) & 89.15\(_{0.13}\) & 89.83\(_{0.07}\) & 88.66\(_{1.19}\) & \textbf{91.58\(_{0.15}\)} & 90.86\(_{0.38}\) & 89.65\(_{0.43}\) \\
    & $\downarrow$CVR   & 0.00\(_{0.00}\) & 100.00\(_{0.00}\) & 18.41\(_{0.91}\) & 100.00\(_{0.00}\) & 89.84\(_{17.59}\) & \textbf{7.34\(_{0.17}\)} & 38.05\(_{0.27}\) \\
    & $\uparrow$Column-wise    & 100.00\(_{0.00}\) & 83.78\(_{1.89}\) & 83.34\(_{5.94}\) & 85.10\(_{0.91}\) & 91.20\(_{0.18}\) & \textbf{98.94\(_{0.13}\)} & 95.12\(_{0.02}\) \\
    & $\uparrow$Pair-wise    & 100.00\(_{0.00}\) & 83.22\(_{2.54}\) & 82.11\(_{8.21}\) & 85.68\(_{1.41}\) & 76.81\(_{3.07}\) & \textbf{97.75\(_{0.52}\)} & 86.73\(_{0.65}\) \\
    \midrule
    \multirow{4}[2]{*}{Default}& $\uparrow$AUC    & 76.48\(_{0.13}\) & 73.50\(_{0.74}\) & 68.67\(_{7.47}\) & 72.55\(_{1.36}\) & 74.29\(_{0.17}\) & \textbf{75.94\(_{0.18}\)} & 74.97\(_{0.78}\) \\
    & $\downarrow$CVR   & 0.00\(_{0.00}\) & 47.44\(_{2.22}\) & 1.22\(_{1.19}\) & 26.46\(_{4.13}\) & 87.78\(_{21.17}\) & 0.71\(_{0.29}\) & \textbf{0.06\(_{0.02}\)} \\
    & $\uparrow$Column-wise    & 100.00\(_{0.00}\) & 86.14\(_{0.39}\) & 85.56\(_{0.66}\) & 88.92\(_{0.33}\) & 93.36\(_{0.04}\) & \textbf{98.35\(_{0.10}\)} & 96.57\(_{0.09}\) \\
    & $\uparrow$Pair-wise    & 100.00\(_{0.00}\) & 76.70\(_{0.83}\) & 82.42\(_{1.55}\) & 80.08\(_{2.75}\) & 27.92\(_{0.93}\) & \textbf{94.13\(_{0.26}\)} & 87.56\(_{1.55}\) \\
    \midrule
    \multirow{4}[2]{*}{shoppers}& $\uparrow$AUC    & 92.60\(_{0.10}\) & 87.95\(_{0.74}\) & 88.08\(_{1.41}\) & 88.48\(_{2.02}\) & \textbf{91.30\(_{0.54}\)} & 81.35\(_{18.36}\) & 88.77\(_{0.76}\) \\
    & $\downarrow$CVR  & 0.00\(_{0.00}\) & 94.72\(_{2.29}\) & 5.99\(_{3.93}\) & 89.20\(_{2.05}\) & 100.00\(_{0.00}\) & 2.44\(_{1.49}\) & \textbf{2.03\(_{0.10}\)} \\
    & $\uparrow$Column-wise    & 100.00\(_{0.00}\) & 77.42\(_{1.78}\) & 70.00\(_{5.41}\) & 76.59\(_{1.23}\) & 83.35\(_{0.01}\) & \textbf{97.11\(_{0.64}\)} & 94.90\(_{0.03}\) \\
    & $\uparrow$Pair-wise    & 100.00\(_{0.00}\) & 86.78\(_{0.21}\) & 70.52\(_{3.27}\) & 81.88\(_{1.44}\) & 52.75\(_{0.10}\) & \textbf{93.87\(_{1.67}\)} & 92.40\(_{0.30}\) \\
    \midrule
    \multirow{4}[2]{*}{Magic}& $\uparrow$AUC    & 94.65\(_{0.21}\) & 82.97\(_{0.61}\) & 87.31\(_{0.45}\) & 90.12\(_{0.47}\) & 88.51\(_{0.59}\) & \textbf{93.21\(_{0.55}\)} & 85.44\(_{0.20}\) \\
    & $\downarrow$CVR   & 0.00\(_{0.00}\) & 12.25\(_{0.99}\) & 2.61\(_{0.80}\) & 1.44\(_{0.22}\) & 55.53\(_{0.06}\) & \textbf{0.73\(_{0.11}\)} & 6.98\(_{0.05}\) \\
    & $\uparrow$Column-wise    & 100.00\(_{0.00}\) & 89.70\(_{2.19}\) & 76.87\(_{0.83}\) & 91.43\(_{0.37}\) & 87.74\(_{0.26}\) & \textbf{98.77\(_{0.19}\)} & 92.31\(_{0.11}\) \\
    & $\uparrow$Pair-wise    & 100.00\(_{0.00}\) & 89.38\(_{0.99}\) & 88.30\(_{0.39}\) & 93.87\(_{0.63}\) & 83.17\(_{0.03}\) & \textbf{99.17\(_{0.23}\)} & 91.10\(_{0.05}\) \\
    \midrule
    \multirow{4}[2]{*}{Beijing}& $\downarrow$RMSE   & 0.42\(_{0.01}\) & 0.85\(_{0.05}\) & 0.96\(_{0.18}\) & 0.83\(_{0.06}\) & 1.32\(_{0.13}\) & 2.17\(_{1.33}\) & \textbf{0.79\(_{0.02}\)} \\
    & $\downarrow$CVR   & 0.00\(_{0.00}\) & 75.25\(_{18.48}\) & 12.16\(_{6.48}\) & 80.08\(_{3.38}\) & 60.98\(_{2.52}\) & 33.11\(_{26.15}\) & \textbf{2.42\(_{0.07}\)} \\
    & $\uparrow$Column-wise    & 100.00\(_{0.00}\) & 80.59\(_{1.18}\) & \textbf{89.30\(_{6.15}\)} & 79.32\(_{1.40}\) & 88.07\(_{0.27}\) & 64.01\(_{30.05}\) & 85.24\(_{0.06}\) \\
    & $\uparrow$Pair-wise   & 100.00\(_{0.00}\) & 80.38\(_{0.62}\) & 77.96\(_{5.60}\) & \textbf{81.67\(_{1.68}\)} & 76.44\(_{2.89}\) & 58.56\(_{31.97}\) & 76.32\(_{0.03}\) \\
    \midrule
    \multirow{4}[2]{*}{News}& $\downarrow$RMSE   & 0.85\(_{0.01}\) & 0.86\(_{0.01}\) & 6.84\(_{0.03}\) & 0.99\(_{0.05}\) & \textbf{0.82\(_{0.01}\)} & 3.10\(_{0.00}\) & 0.85\(_{0.02}\) \\
    & $\downarrow$CVR   & 0.00\(_{0.00}\) & 100.00\(_{0.00}\) & 100.00\(_{0.00}\) & 100.00\(_{0.00}\) & 100.00\(_{0.00}\) & 100.00\(_{0.00}\) & \textbf{99.24\(_{0.04}\)} \\
    & $\uparrow$Column-wise   & 100.00\(_{0.00}\) & 83.86\(_{0.66}\) & 49.56\(_{1.74}\) & 83.07\(_{0.39}\) & \textbf{96.98\(_{0.03}\)} & 13.52\(_{6.21}\) & 96.45\(_{0.07}\) \\
    & $\uparrow$Pair-wise    & 100.00\(_{0.00}\) & 94.82\(_{0.09}\) & 89.04\(_{0.43}\) & 93.41\(_{0.03}\) & 88.49\(_{0.10}\) & 28.45\(_{47.59}\) & \textbf{98.35\(_{0.03}\)} \\
    \bottomrule
  \end{tabular}%
  \label{tab:result}%
\end{table*}%
\subsection{Privacy Protection} Another important objective of synthetic tabular data is to enable data generation in privacy-sensitive domains, such as finance and healthcare, without disclosing sensitive information that could potentially identify individuals from the original dataset. Therefore, it is essential to evaluate the synthetic data using privacy protection metrics to ensure that privacy protection is achieved. These metrics aim to quantify the extent to which the records in the real dataset can be re-identified or inferred from the synthetic data. They are designed to assess whether the synthetic data inadvertently leaks sensitive information, either through direct resemblance to individual records or through indirect statistical inference. The primary privacy protection evaluation techniques include:
\begin{itemize}
    \item \textbf{Distance to Closest Record (DCR):} This metric measures the minimum distance between each synthetic record and its nearest neighbor in the real dataset. A small DCR value may indicate that some synthetic samples are too similar to real individuals, raising concerns about memorization and potential privacy leakage. Typically, larger average DCR values are preferred, as they imply that synthetic records are sufficiently distinct from real ones.
    \item \textbf{Attribute Inference Attack:} This type of privacy attack assumes that an adversary has partial knowledge of an individual’s data and attempts to infer missing or sensitive attributes using the synthetic tabular dataset \( D_s \). The success of such an attack reflects the extent to which private information can be predicted from synthetic data. High inference accuracy indicates poor privacy protection and potential leakage of attribute-level information.
    \item \textbf{Membership Inference Attack:} In this setting, an attacker tries to determine whether a specific individual was part of the training data used to generate the synthetic tabular dataset \( D_s \). This is typically done by exploiting differences in model behavior on seen versus unseen data. A high success rate in membership inference suggests that the generative model may have overfit the training data, leading to a privacy protection risk.
\end{itemize}
Together, these metrics provide a comprehensive view of the privacy risks associated with synthetic tabular data and are crucial for validating the safety of data release in sensitive application domains.

\subsection{Analysis and Discussion} To intuitively demonstrate the performance of different synthetic tabular data generation methods, we select a representative model from each summarized category and conduct experiments on several classic datasets. The experimental results are then analyzed and discussed. The results are shown in Table~\ref{tab:result}. Compared to traditional generation methods, tabular data produced by Diffusion Model methods and LLM-based approaches exhibits markedly higher data availability. Nonetheless, the hallucination problem associated with LLMs poses a challenge to fidelity, especially in terms of pairwise fidelity. Diffusion models exhibit superior performance in generating synthetic tabular data with a small number of columns, while LLM-based methods are more effective in handling tabular data with a greater number of columns. While the diffusion model excels in processing numerical features, its performance advantage diminishes on datasets characterized by a higher proportion of categorical features, such as News and Shoppers.
\section{Practical Applications and Future Directions}
Given its importance in data augmentation and privacy protection, synthetic tabular data generation has gained increasing attention in various practical scenarios. This section presents representative real-world applications and outlines promising avenues for future research.
\subsection{Practical Applications}
In this section, we examine the practical impact of synthetic tabular data generation from two distinct yet complementary perspectives: its use in improving tabular data availability and its application in privacy protection data analysis.

\subsubsection{Improving Tabular Data Availability}
In practical applications, tabular data often suffers from class imbalance and missing values. These challenges can significantly impair the performance of downstream machine-learning tasks. As a result, improving data availability through the use of synthetic data has emerged as an essential strategy to alleviate such limitations and improve model robustness. \textit{For cases of imbalanced categories}, models such as cGAN~\cite{douzas2018effective}, CTAB-GAN~\cite{zhao2021ctab}, and CTGAN~\cite{xu2019modeling} incorporate conditional vectors into the GAN framework to guide the generation process and specifically synthesize samples from minority classes. SOS~\cite{kim2022sos} proposes a style-transfer-based oversampling method for handling imbalanced tabular data using score-based generative models (SGMs). The core idea is to transform the majority-class samples into minority-class samples through a score-based diffusion process, thereby addressing class imbalance in a more targeted and data-efficient manner. EPIC~\cite{kim2024epic} and LITO~\cite{yang2024language} leverage prompt engineering to instruct LLMs to generate synthetic samples from minority classes. By crafting task-specific prompts, these methods effectively address class imbalance in tabular datasets through controlled, LLM-driven data generation. \textit{For the case of missing values}, one common approach for handling missing values is Multiple Imputation by Chained Equations (MICE)~\cite{white2011multiple}, an iterative method that estimates the conditional distribution of each feature given the others and imputes missing values accordingly. 
Diffusion Model Methods~\cite{shi2024tabdiff,lin2024ctsyn} typically generates missing values by conditioning on the column names of the missing entries. LLMs that have been fine-tuned on tabular data possess an inherent ability to infer missing values by leveraging prompt-based generation, without the need for retraining or explicit imputation models~\cite{kim2024epic,seedat2023curated,borisov2022language}.

\subsubsection{Privacy Protection}
Additionally, in domains such as finance and healthcare, strict compliance with data privacy protection regulations is required to ensure the security and confidentiality of user data. As a result, the use of synthetic data for privacy protection has become particularly important in these sensitive application areas. medGAN~\cite{choi2017generating} and medBGAN~\cite{baowaly2019synthesizing} apply generative adversarial networks (GANs) to the synthesis of electronic health records (EHRs). The effectiveness of the generated synthetic tabular data has been demonstrated through a combination of statistical distribution comparisons, predictive modeling tasks, and expert evaluations by medical professionals. These studies also show that synthetic data poses limited risks in terms of identity and attribute disclosure, thereby supporting its utility in privacy-sensitive healthcare applications. EHRDiff~\cite{yuan2023ehrdiff}, TabDDPM-EHR~\cite{ceritli2023synthesizing}, MedDiff~\cite{he2023meddiff}, and DPM-EHR~\cite{nicholas2023synthetic} explore the use of diffusion models for synthetic tabular data generation in order to produce more realistic and high-fidelity synthetic electronic health records (EHRs). These methods leverage the denoising capabilities and distributional flexibility of diffusion-based generative frameworks to better capture the complex patterns inherent in medical tabular data. FLEXGEN-EHR~\cite{he2024flexible} further advances this line of research by explicitly addressing the challenge of missing modalities in EHRs. It introduces a unified learning framework that aligns heterogeneous data representations and effectively imputes missing information, thereby enhancing the completeness and utility of the generated synthetic records.
\subsection{Future Directions}
Synthetic tabular data generation is a rapidly evolving field that remains in its early stages. The field continues to face several unresolved challenges, which warrant further investigation in future research:
\begin{itemize}
    \item \textbf{Trade-off Between Data Availability and Privacy Protection.} Many early methods have prioritized either data utility or privacy protection, often resulting in highly specialized models with limited generalizability. This highlights a key challenge in the field: achieving a balanced trade-off between data utility and privacy protection.
    \item \textbf{Alignment with human knowledge.} Current generative models typically learn the distribution of the original data and generate synthetic samples that conform to it. However, distributional alignment does not necessarily guarantee consistency with human commonsense knowledge. This misalignment can introduce risks in data usage, particularly as generative models are increasingly deployed in real-world applications. With the growing emphasis on neuro-symbolic AI and safe AI, ensuring that synthetic tabular data aligns with human knowledge and domain-specific logic is becoming an increasingly important research direction.
    \item \textbf{Enhancing interpretability}. Most existing generative models operate as black-box systems, making it difficult to understand their internal decision-making processes. As a result, these models are susceptible to inheriting and amplifying biases present in the original data, potentially leading to the generation of biased synthetic tabular datasets \( D_s \). Therefore, developing methods to improve model interpretability is critical for ensuring the transparency, fairness, and trustworthiness of synthetic tabular data generation, especially in high-stakes applications.
    \item \textbf{Evaluation metrics.} Unlike image generation, where visual quality often serves as a direct and intuitive evaluation criterion, assessing the quality of synthetic tabular data is significantly more complex. It frequently requires domain-specific metrics, particularly in sensitive fields such as healthcare and finance, where the correctness, consistency, and utility of the data must be rigorously validated against real-world constraints and expert knowledge.
\end{itemize}

\section{Conclusion}
In this work, we present a comprehensive survey of synthetic tabular data generation, covering the end-to-end generation process and introducing a novel taxonomy to categorize existing generation methods. Our goal is to provide practical guidance for enterprises and organizations seeking to adopt synthetic tabular data generation techniques. At the same time, we offer insights into the key challenges and opportunities within the field, along with a thorough summary of evaluation protocols and real-world applications. Furthermore, we highlight promising directions for future research. We hope that this work will contribute to advancing the development of synthetic tabular data generation and inspire further exploration in this rapidly evolving area.

 
\bibliographystyle{IEEEtran}
\bibliography{IEEEtran}

\end{document}